\documentclass[conference]{IEEEtran}
\IEEEoverridecommandlockouts
\usepackage{cite}
\usepackage{amsmath,amssymb,amsfonts}
\usepackage{algorithmic}
\usepackage{graphicx}
\usepackage{textcomp}
\usepackage{xcolor}
\usepackage{url}
\usepackage{hyperref}
\def\BibTeX{{\rm B\kern-.05em{\sc i\kern-.025em b}\kern-.08em
    T\kern-.1667em\lower.7ex\hbox{E}\kern-.125emX}}
    
\begin{document}

\title{Who killed Lilly Kane? A case study in applying knowledge graphs to crime fiction.
\thanks{We are grateful for support from UCLA and Harvey Mudd College as well as the Los Angeles City College STEM Pathways program, supported by Department of Education PR\# P031C160251.
This work was also partially supported by NSF grants DMS-1737770 and DMS-2027277.}
}

\author{\IEEEauthorblockN{Mariam Alaverdian}
\IEEEauthorblockA{\textit{Department of Mathematics} \\
\textit{Los Angeles City College}\\
Los Angeles, CA 90029 \\
masha.alaverdyan@gmail.com}
\and
\IEEEauthorblockN{William Gilroy}
\IEEEauthorblockA{\textit{Department of Mathematics} \\
\textit{Harvey Mudd College}\\
Claremont, CA \\
wgilroy@hmc.edu}
\and
\IEEEauthorblockN{Veronica Kirgios}
\IEEEauthorblockA{\textit{Department of Mathematics} \\
\textit{Notre Dame Univ.}\\
Notre Dame, IN 46556 \\
vkirgios@nd.edu}
\and
\IEEEauthorblockN{Xia Li}
\IEEEauthorblockA{\textit{Department of Mathematics} \\
\textit{University of California, Los Angeles}\\
Los Angeles, USA 90095\\
xli51@math.ucla.edu}
\and
\IEEEauthorblockN{Carolina Matuk}
\IEEEauthorblockA{\textit{Department of Mathematics} \\
\textit{Univ. of Iowa}\\
Iowa City, Iowa 52242-1419 \\
carolinamatuk@hotmail.com}
\and
\IEEEauthorblockN{Daniel McKenzie}
\IEEEauthorblockA{\textit{Department of Mathematics} \\
\textit{University of California, Los Angeles}\\
Los Angeles, USA 90095\\
mckenzie@math.ucla.edu}
\and
\IEEEauthorblockN{Tachin Ruangkriengsin}
\IEEEauthorblockA{\textit{Department of Mathematics} \\
\textit{University of California, Los Angeles}\\
Los Angeles, USA 90095\\
bankcrub@g.ucla.edu}
\and
\IEEEauthorblockN{Andrea L. Bertozzi}
\IEEEauthorblockA{\textit{Department of Mathematics} \\
\textit{University of California, Los Angeles}\\
Los Angeles, USA 90095\\
bertozzi@ucla.edu}
\and
\IEEEauthorblockN{P. Jeffrey Brantingham}
\IEEEauthorblockA{\textit{Department of Anthropology} \\
\textit{University of California, Los Angeles}\\
Los Angeles, USA 90095\\
branting@ucla.edu}
}

\maketitle

\begin{abstract}
We present a preliminary study of a knowledge graph created from season one of the television show {\em Veronica Mars}, which follows the eponymous young private investigator as she attempts to solve the murder of her best friend Lilly Kane. We discuss various techniques for mining the knowledge graph for clues and potential suspects. We also discuss best practice for collaboratively constructing knowledge graphs from television shows.
\end{abstract}


\section{Introduction}
Knowledge graphs are a powerful tool for organizing, storing and presenting complex data. A knowledge graph is a graph of data whose nodes represent entities of interest and whose edges represent relations between these entities. Formally, a knowledge graph consists of a set of entities, $\mathcal{V}$, a set of relations or predicates, $\mathcal{R}$ and a set of facts, $\mathcal{E}$, which specify pairwise relations between entities. Crucially, the facts must obey rules specified by an accompanying ontology, $\mathcal{O}$, which dictates which kinds of relations can be present between which kinds of entities. For a comprehensive and modern introduction to knowledge graphs, we refer the reader to \cite{hogan2020knowledge}. 

\subsection{Knowledge graphs for fiction}
Despite the promise that knowledge graphs hold in analyzing semantic data, they have not yet been extensively applied in analyzing fiction. We believe that knowledge graphs are a promising tool for encoding and analyzing the complex human-human interactions present in novels, movies and television shows, and further that these domains form a realistic proxy for real-world human-human interactions. Inspired by a knowledge graph challenge problem \cite{kawamura2019report}, we chose to study the genre of crime fiction. Unlike the Sherlock Holmes novel studied in \cite{kawamura2019report}, we chose to focus on a television show, {\em Veronica Mars}. While there has been prior work on analyzing movies and television shows using graph theory \cite{bonato2016mining, bonato2018dynamic}, we believe that we are the first to apply {\em knowledge graphs} to television. Television has at least two distinct advantages over novels:
\begin{itemize}
    \item T.V. scripts are more structured than novels. In principle this makes it easier to automate knowledge graph construction.
    \item Continuity over episodes and seasons allows for the construction of a larger and richer knowledge graph.
\end{itemize}

\subsection{Veronica Mars}
Set in a fictional town in California, {\em Veronica Mars} is a modern day spin on {\em Nancy Drew}. The teenage protagonist, Veronica Mars, is a private investigator. The series begins with the murder of Veronica's best friend, Lilly Kane. A minor character, Abel Koontz, is convicted of her muder but there is reason to suspect that he is not the true culprit. This search for Lilly's true murderer forms the overarching plot motif. Within each episode Veronica is presented with a crime, gathers evidence and (usually) solves the case. Unlike popular detective fiction such as Sherlock Holmes, the facts of the case are usually transparently presented throughout the episode and the identity of the culprit can usually be deduced straightforwardly from these facts, without recourse to {\em deus ex machina}.

\subsection{Notation}
We shall denote our knowledge graph as $\mathcal{G} = (\mathcal{V},\mathcal{R}, \mathcal{E})$. Entities and relations shall be written in typewriter font, for example {\tt VeronicaMars} \footnote{and note the distinction between the fictional person, Veronica Mars, the television show, {\em Veronica Mars}, and the entity in our knowledge graph {\tt Veronica\_Mars}!}. Facts shall be written as triples of the form ({\tt Entity1},{\tt Relation},{\tt Entity2}). Occasionally (see Section \ref{sec:Link_Prediction}), we shall suppose partial knowledge of $\mathcal{G}$, and in particular assume that there are true relations between entities that are not captured by $\mathcal{E}$. We shall refer to these unknown triples as unknown or missing facts. We shall denote by $G$ the underlying undirected graph of $\mathcal{G}$.

We experimented with various software for constructing knowledge graphs, for example Karma \cite{knoblock2012semi}. However we found that our data was too unstructured for most such tools. The process of constructing a knowledge graph from {\em unstructured} data is less well studied, although we note the recent works \cite{liu2018seq2rdf,kertkeidkachorn2017t2kg}. In particular, Seq2RDF \cite{liu2018seq2rdf} is a powerful tool for turning sentences into triples that agree with a given ontology. In principle such a tool could be used to automate the construction of a knowledge graph from the script of a T.V. show. However, we found several roadblocks to this approach:
\begin{itemize}
    \item State-of-the-art tools such as Seq2RDF are currently only able to extract one triple per sentence, and these triples need to be of a fairly quotidian nature (for example ``Berlin is the capital of Germany''). They are not able to handle compound sentences, synonyms or metaphors which are common in fiction.
    \item Some important facts are conveyed over multiple, non-contiguous sentences, conveyed visually or implied without ever being explicitly stated.
\end{itemize}
Thus, we chose to manually construct the knowledge graph. Our workflow was as follows: two or three team members would watch an episode and then collaboratively identify which facts presented in the episode were important. These were then captured in a spreadsheet. The spreadsheets for each episode were then combined, and the Python package RDF lib was used to convert them into a collection of RDF triples. Further details are presented in Sections \ref{sec:Ontology} and \ref{sec:WhichFacts}. Our resulting knowledge graph contains 541 entities and 1,106 facts. The complete data set, as well as all Python code used in constructing it, is available at \cite{githublink}. 

\subsection{The ontology and allowed relations}
\label{sec:Ontology}
\begin{figure}[h!]
    \centering
    \includegraphics[width = 0.8\linewidth]{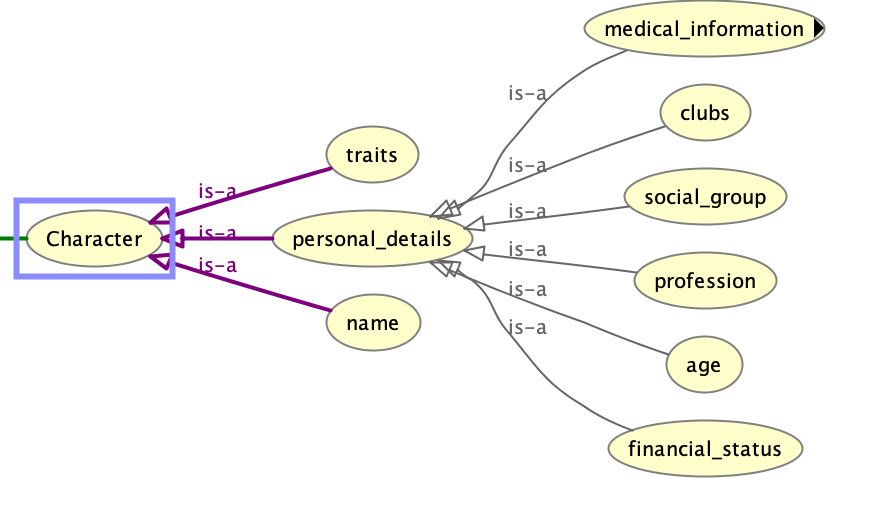}
    \caption[]{A snapshot of hierarchical relations in the ontology.}
    \label{fig:Ontology}
\end{figure}

Our ontology was based on the ``friend of a friend'' ontology, and thus allows for common relations between characters (e.g. ``friend of'', ``child of''). The ontology automatically encodes the fact that some relations are symmetric (if ``A is friend of B'' then also ``B is friend of A'') or have obvious inverses (if ``A is child of B'' then ``B is parent of A''). We augmented this ontology to allow for relations between characters particular to crime fiction (``A is kidnapped by B'') as well as relevant relations between characters and places, objects and abstract concepts such as financial status. We took a flexible approach and added new relations as they occurred in the show. A representative portion of our ontology is displayed as Figure \ref{fig:Ontology}. 

\subsection{Which facts are important?}
\label{sec:WhichFacts}
The process of choosing which facts in a given episode to record and add to the knowledge graph is somewhat subjective. We focused on facts that:
\begin{itemize}
\item Captured important biographical information of characters.
\item Captured important relationships between characters.
\item May be clues, both for the episode specific case and the overarching case.
\end{itemize}
For example, the case in episode six centers on a rigged student council election, yielding triples such as ({\tt Wanda\_Varner}, {\tt runs\_for}, {\tt student\_council}) and ({\tt Madison\_Sinclair}, {\tt threatens}, {\tt Wanda\_Varner}). However, key clues to the overarching case ({\em i.e.} the murder of Lilly Kane) are also revealed in this episode, yielding: ({\tt white\_sneakers}, {\tt clue\_of}, {\tt Case1}) and ({\tt white\_sneakers}, {	\tt seen\_at}, {\tt Lilly\_Kane's\_room}). More examples may be found in Table \ref{tab:Sample_Facts}.

\begin{table}[h!]
    \centering
    \begin{tabular}{c|c|c}
    Subject & Predicate & Object \\
    \hline
    {\tt Veronica\_Mars} & {\tt child\_of} & {\tt Keith\_Mars}\\
    {\tt Weevil\_Navarro} & {\tt has\_financial\_status} & {\tt lower\_class}\\
    {\tt Wallace\_Fennel} & {\tt in\_club} & {\tt basketball}\\
    {\tt Don\_Lamb} & {\tt employee\_of} & {\tt Keith\_Mars}\\
    {\tt Van\_Clemmons} & {\tt has\_last\_name} & {\tt Clemmons}\\
    \hline
    \end{tabular}
    \caption{Some character-specific facts in the {\em Veronica Mars} knowledge graph}
    \label{tab:Sample_Facts}
\end{table}

\subsection{Reification}
In analyzing crime fiction, recording the time at which certain facts are revealed is essential. To do this we applied reification, a technique originally used for reifying an \emph{attributed knowledge graph}, to record temporal information. In the process of reification, relations are first ``reified'' into entities. These entities can then be the subject of a fact. We then add the relation {\tt occurs\_at} to $\mathcal{R}$. A time stamp can then be added by including the fact ({\tt Reified\_relation},{\tt occurs\_at}, time\_stamp) in $\mathcal{G}$. \\

We note that reification could also be used to record the {\em provenance} of facts. For example, if the fact ``character-A was seen at location-B'' is revealed by character-C, it could be useful to record this. Thus, if at a later stage character-C is revealed to be unreliable, one can easily search for the facts that may no longer be valid.

\subsection{Character subgraphs}
\begin{figure}
    \centering
    \includegraphics[width = 0.5\textwidth]{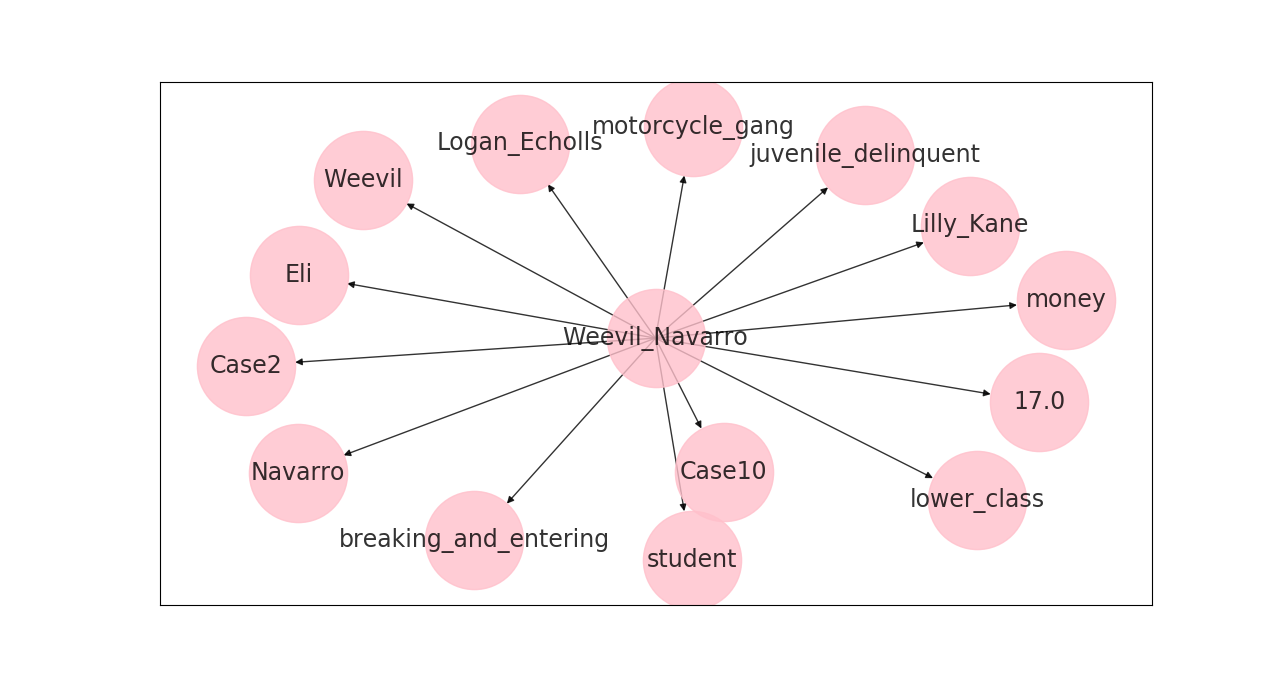}
    \caption{The character subgraph for Weevil Navarro, with temporal labels suppressed.}
    \label{fig:Weevil}
\end{figure}

Because the knowledge graph is constructed using information from every episode, one can easily query it to produce longitudinal subgraphs. For example, one can extract the subgraph of all triples containing a particular character as a subject. This provides a snapshot of a character's story arc through the series, and also identifies the cases they were involved in. See, for example, Figure \ref{fig:Weevil}.

\section{Analysis}
With our knowledge graph in hand, we attempted to use it to answer several questions:
\begin{enumerate}
    \item Can we identify important clues for a given case?
    \item Can we extract overarching themes or topics from the knowledge graph that might not be apparent given only individual episodes?
    \item Can we identify who killed Lilly Kane?
\end{enumerate}
In this section we present preliminary results on all three of these problems.

\subsection{Identifying relevant clues}
\begin{figure}
    \centering
    \includegraphics[width = 0.4\textwidth]{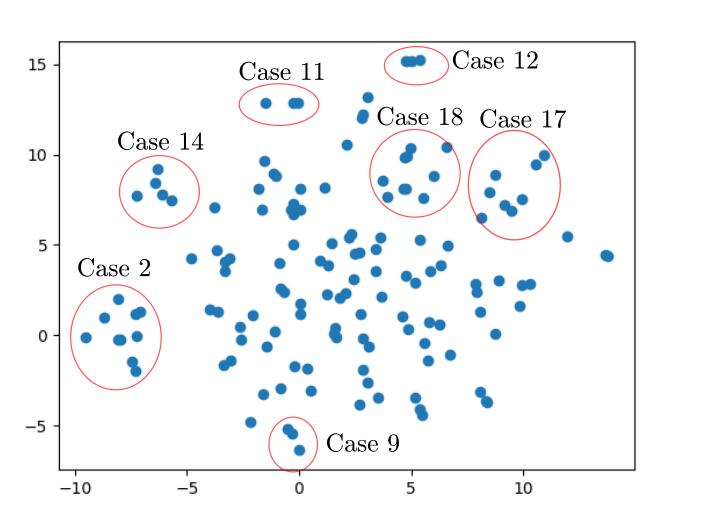}
    \caption{2D TSNE representation of the {\tt TransE} embedding. Here we have represented the embeddings of clues from various cases. The red circles indicate that clues which are from the same case, in other words semantically similar entities, seem close to each other in the TransE emebedding. For example, in the cluster of case 11, the entities consist of the {\tt asphyxiation} (the means of the crimes), {\tt Vic Sciaraffa} (the character), {\tt videotapes}, {\tt number} and { \tt wristband} (the clues).}
    \label{fig:TransE}
\end{figure}

We used TransE \cite{bordes2013translating} to embed our knowledge graph into $\mathbb{R}^{200}$. Recall that TransE assigns every entity $v\in\mathcal{V}$ to a vector $\mathbf{u}_{v}\in\mathbb{R}^{200}$ {\em and} every relation $r\in\mathcal{R}$ to a vector $\mathbf{u}_{r}\in\mathbb{R}^{200}$ such that, for every fact $(v_1,r,v_2)\in\mathcal{E}$ we have that $\mathbf{u}_{v_1} + \mathbf{u}_{r} \approx \mathbf{u}_{v_2}$. Specifically, TransE aims to find an embedding minimizing the loss:
$$
\sum_{(v_1,r,v_2)\in\mathcal{E}} \|\mathbf{u}_{v_1} + \mathbf{u}_{r} - \mathbf{u}_{v_2}\|_2 - \sum_{(v_1,r,v_2)\not\in\mathcal{E}} \|\mathbf{u}_{v_1} + \mathbf{u}_{r} - \mathbf{u}_{v_2}\|_2
$$
TransE is designed to place semantically similar entities and relations together, so we hypothesized that it would place relevant clues close to the case with which they were associated. To visualize this embedding, we used t-SNE \cite{maaten2008visualizing} to project from $\mathbb{R}^{200}$ to $\mathbb{R}^{2}$ (see \cite{githublink} for details). The result of this projection is shown in Figure \ref{fig:TransE}. Even after this enormous reduction in dimension, we observe some encouraging results. For at least $7$ of the $19$ cases, multiple relevant clues for these cases are indeed grouped together (shown circled in red).

\subsection{Topic extraction using random walks}


Given a corpus $\{d_1,\ldots, d_n\}$ of documents, the process of extracting relevant topics from this corpus is well studied:
\begin{enumerate}
    \item Identify $m$ keywords relevant to this corpus.
    \item For each $d_i$ construct a vector $\mathbf{v}_i$ of (possibly weighted) counts of keywords.
    \item Form the matrix $X = \left[\begin{matrix} \mathbf{v}_1 & \ldots & \mathbf{v}_n\end{matrix}\right]$.
    \item Compute a (low-rank) non-negative matrix factorization: $X \approx UV$ with $U\in\mathbb{R}^{m\times r}$ and $V\in\mathbb{R}^{r\times n}$
    \item Obtain $r$ topics, one for each column of $U$.
\end{enumerate}
The implicit assumption that makes this procedure work is that the number of topics is much less than the number of documents, and that each document is a superposition of a small number of topics.\\

It is not clear how to adapt this to analyze a season of a television show. The key issue is how to ``slice'' the season into documents. The naive solution of declaring each episode to be a document is unsatisfactory because:
\begin{itemize}
    \item There are relatively few episodes in a season.
    \item A given story arc is often developed over multiple episodes. For example, in {\em Veronica Mars} details relating to Lilly Kane's murder are presented to the viewer in the form of flashbacks interspersed throughout season one.
\end{itemize}
Inspired by \cite{lyu2019online,lyu2019sampling} we propose a novel approach:
\begin{enumerate}
    \item Fix a required number of documents, $n$. 
    \item Let $G$ denote the undirected graph obtained by forgetting the orientation and type of each relation in $\mathcal{G}$.
    \item For $i=1,\ldots, n$, choose a random initial vertex $v_{0}^{(i)}$ and perform a $\ell$ step random walk starting from $v_{0}^{(i)}$. Let:
    $$
   d_{i} =  \left(v_{0}^{(i)}, e_{1}^{(i)},v_{1}^{(i)},\ldots, v_{\ell}^{(i)} \right)
    $$
    be the collection of vertices and edges traversed by this walk.
    \item Using $d_{1},\ldots, d_{n}$ our documents, perform the topic modeling as described earlier.\\
\end{enumerate}

Note that using this ``motif sampling'', one can easily generate a large corpus from a single season. More importantly, we hypothesize that these randomly sampled ``motifs'' are more likely to capture important topics than episodes, as they will contain related elements that span multiple episodes. We applied the technique to the {\em Veronica Mars} knowledge graph with $n = 1000$ and $l = 50$. We used TF-IDF to vectorize the resulting documents, and performed NMF with $r = 25$. The reason we use TF-IDF here is to balance out the importance of nodes---as nearly every character is related to Veronica, it is very likely that every random walk will contain {\tt Veronica\_Mars}. While the resulting topics were somewhat noisy, the results are encouraging. For example, Topic 12 contains the entities {\tt Aaron\_Echolls}, {\tt tapes} {\tt affair\_with} and {\tt Lilly\_Kane}. This topic neatly summarizes the circumstances of Lilly Kane's death: Aaron Echolls killed Lilly Kane in a fit of rage after video tapes documenting their affair came to light. Other topics relate to important characters (Topic 13) or to episode-specific cases (Topic 20). The complete list of topics is available at \cite{githublink}.

\begin{table}[h!]
    \centering
    \begin{tabular}{c|c|c}
         Topic 12 & Topic 13 & Topic 20 \\
         \hline
         {\tt Aaron\_Echolls} & {\tt Duncan\_Kane} & {\tt Wanda\_Varner} \\
         {\tt Lilly\_Kane} &{\tt blackout} & {\tt rigged\_election} \\
         {\tt affair\_with} &{\tt epilepsy} &  {\tt ballot\_instructions} \\
         {\tt tapes} & {\tt oxcarbazepine} & {\tt Madison\_Sinclair}\\
    \end{tabular}
    \caption{Selected topics}
    \label{tab:my_label}
\end{table}

\subsection{Link Prediction}
\label{sec:Link_Prediction}
The link prediction problem takes two entities, $v_1$ and $v_2$, and a relation $r\in\mathcal{R}$ and asks whether the triple $(v_1,r,v_2)$ should be a fact in the knowledge graph $\mathcal{G}$. Ideally, one would like to use link prediction to deduce the guilty parties in the various cases solved by Veronica Mars by taking $v_1=${\tt Character\_A}, $r=${\tt described\_as} and $v_2=$ {\tt perpetrator}. Due to the lack of training data (there are only 20 cases), we found this challenging. Thus, we also investigated using link prediction to determine whether or not two characters were friends by choosing $v_1=${\tt Character\_A}, $v_2=$ {\tt Character\_B} and $r=${\tt friend\_of}.\\

Many approaches to link prediction first construct a vector embedding of $\mathcal{G}$ and then assign a probability to $(v_1,r,v_2)$ being a fact inversely proportional to $\|\mathbf{u}_1 + \mathbf{u}_{r} - \mathbf{u}_{v_2}\|_2$ (see \cite{rossi2020knowledge} for an overview). We experimented with using TransE for link prediction in this manner, but found the results to be unsatisfactory. We hypothesize that this is because the complex social links we are seeking to predict are more appropriately captured by a subgraph than by an embedding. For example, the data of a crime could be represented by a subgraph connecting the perpetrator, the victim, a motive for the crime, a location of the crime and several damning pieces of evidence.\\

Motivated by this hypothesis, we investigated link prediction algorithms that are subgraph based. In particular, we used SEAL \cite{zhang2018link}. Given a putative triple $(v_1,r,v_2)$, SEAL constructs an enclosing subgraph around this link and then uses a trained graph neural network (GNN) to output a probability of this link being a true fact. It is important to note that SEAL is designed for undirected graphs, and thus is not cognizant of the type or direction of the relation $r$. SEAL also does not use the ontology in any way. Like any GNN, SEAL requires training in the form of positive examples of the link we are trying to predict. Negative examples are generated by random sampling. We experimented with 50\%, 75\% and 90\% of our data as training data (with the rest held back as a test set). \\

We found mixed results. For example, using a 90\% split SEAL assigned an encouraging $71\%$ probability to the triple $({\tt Aaron\_Echolls},{\tt described\_as},{\tt perpetrator})$, correctly identifying Aaron Echolls as a suspect even though this was not given in the training data. Unfortunately it assigned similarly high probabilities to false triples such as $({\tt Logan\_Echolls},{\tt described\_as},{\tt perpetrator})$. 
We suspect that the poor performance of SEAL here can be explained by the paucity of training data---our knowledge graph has only 541 vertices, whilst the examples considered in \cite{zhang2018link} all have at least 10,000. Moreover, SEAL ignores the rich information encoded in the {\em types} of relations, treating them all uniformly as edges. A recent work \cite{teru2019inductive} extends SEAL to handle multiple kinds of relations and might yield better results.

\section{Conclusions and future directions}

In this paper we considered the novel problem of studying television series using knowledge graphs. We introduced a new knowledge graph data set, which may be of interest to the community. We proposed several novel analysis techniques, such as random walk topic modelling, and tested the applicability of existing techniques to this new domain. We believe that applying knowledge graphs to fiction has tremendous scope, and for future groups we offer the following recommendations:
\begin{itemize}
    \item Before adding any facts to the knowledge graph, develop an application-specific ontology allowing for fewer relations. When developing the knowledge graph, only add additional relations if strictly necessary. This will mitigate a problem that we encountered, namely rare relations that only occur once or twice in the knowledge graph.
    \item While it is tempting to use natural language processing to automate the process of extracting facts, we found these tools unable to deal with the linguistic complexity of fiction. Hence, we recommend manually extracting facts.
    \item We recommend using time stamps. This can be done either using reification (as we have done) or by using an attributed knowledge graph.
    \item The proposed random walk topic modelling scheme seems very promising. An interesting question for future groups would be to compute statistics on coverage ({\em i.e.} how many entities are included in at least one document) and repetition ({\em i.e.} how many documents the average entity appears in). This could assist in selecting the number of documents to generate from a given a knowledge graph. Another interesting line of research would be to make the random walk {\em ontology aware}. For example, if the edge traversed at step $i$ represents a certain kind of relation, this information could be used to restrict the relations which the edges at step $i+1$ can represent. It seems likely that incorporating such additional structural information into the random walks will yield more coherent documents. 
    \item We suspect that GNN approaches may play an increasingly important role in link prediction; further studies might consider starting their work with GRAIL \cite{teru2019inductive}.
    \item Template matching \cite{moorman2018filtering,kawamura2019report} is promising strategy for identifying meaningful subgraphs within the knowledge graph. We note that this approach would benefit from a more principled ontology with fewer relations, as discussed above.
\end{itemize}

\bibliographystyle{alpha}
\bibliography{Bib.bib}
\end{document}